\documentclass{article}

\usepackage[final]{neurips}

\usepackage{enumitem}
\usepackage{marvosym} 
\usepackage[utf8]{inputenc} 
\usepackage[T1]{fontenc}    
\usepackage{hyperref}       
\usepackage{url}            
\usepackage{booktabs}       
\usepackage{amsfonts}       
\usepackage{microtype}      
\usepackage{xcolor}         
\usepackage{epigraph}
\usepackage[export]{adjustbox}
\usepackage{multirow}
\usepackage{listings}
\usepackage{colortbl}
\usepackage{color}
\usepackage{pifont}
\usepackage{fontawesome}
\usepackage{wrapfig}
\usepackage{subcaption}
\usepackage{caption}
\usepackage{float}
\usepackage{enumitem}
\usepackage{tabularx}
\usepackage{makecell}
\usepackage{minitoc} 

\noptcrule

\usepackage{array}
\renewcommand{\arraystretch}{1.2}

\usepackage{caption}
\usepackage{amsmath,amsfonts,bm}

\def\eqref#1{equation~\ref{#1}}

\def\1{\bm{1}}

\DeclareMathAlphabet{\mathsfit}{\encodingdefault}{\sfdefault}{m}{sl}
\SetMathAlphabet{\mathsfit}{bold}{\encodingdefault}{\sfdefault}{bx}{n}

\usepackage{xspace}

\newcommand{\model}[1]{\textsc{Rho-1}} 
\newcommand{\method}[1]{{SLM}} 
\newcommand{\methodfull}[1]{{Selective Language Modeling}} 

\newcommand{\mamix}[1]{\textsc{MaMix}}
\newcommand{\genmix}[1]{\textsc{GenMix}}

\makeatletter  
\renewcommand\@fnsymbol[1]{%
  \ensuremath{%
    \ifcase#1\or 
    \star\or     
    \diamond\or   
    \dagger\or  
    \mathsection\or 
    \mathparagraph\or 
    \|\or        
    **\or        
    \dagger\dagger\or 
    \ddagger\ddagger 
    \else \@ctrerr \fi%
  }%
}  
\makeatother

\definecolor{darkblue}{rgb}{0, 0, 0.5}
\hypersetup{colorlinks=true, citecolor=darkblue, linkcolor=darkblue, urlcolor=darkblue}
\definecolor{lightgray}{rgb}{0.9, 0.9, 0.9}

\definecolor{darkgreen}{RGB}{50,100,0}
\definecolor{darkred}{RGB}{200, 0, 0}
\definecolor{lightred}{RGB}{250, 200, 200}
\definecolor{lightblue}{RGB}{210, 220, 250}
\definecolor{doderblue}{RGB}{30,144,255}
\definecolor{select}{RGB}{222, 235, 247}
\definecolor{unselect}{RGB}{247, 207, 206}

\usepackage{cleveref}
\makeatletter
\AtBeginDocument{%
  \renewcommand{\sectionautorefname}{\S\@gobble}

  \renewcommand{\subsectionautorefname}{\S\@gobble}  
}
\makeatother

\title{Step-Video-TI2V Technical Report: A State-of-the-Art Text-Driven Image-to-Video Generation Model}

\author{Step-Video Team
\\
StepFun
}

\begin{document}

\doparttoc
\faketableofcontents

\maketitle

\begin{abstract}
\label{sec:abstract}
\vspace{-0.2cm}

We present \textbf{Step-Video-TI2V}, a state-of-the-art text-driven image-to-video generation model with 30B parameters, capable of generating videos up to 102 frames based on both text and image inputs.
We build \textbf{Step-Video-TI2V-Eval} as a new benchmark for the text-driven image-to-video task and compare Step-Video-TI2V with open-source and commercial TI2V engines using this dataset. Experimental results demonstrate the state-of-the-art performance of Step-Video-TI2V in the image-to-video generation task. 
Both Step-Video-TI2V and Step-Video-TI2V-Eval are available at \hyperlink{https://github.com/stepfun-ai/Step-Video-TI2V}{https://github.com/stepfun-ai/Step-Video-TI2V}.
\end{abstract}
\section{Introduction}

Text-driven image-to-video (TI2V) models are favored by video content creators because they offer greater control compared to text-to-video (T2V) models. Existing commercial video generation engines like Gen-3 \cite{runwaygen3}, Kling \cite{kling} and Hailuo \cite{hailuo} offer this capability to users. Recently, open-source TI2V models like HunyuanVideo-I2V \cite{kong2024hunyuanvideo} and Wan2.1 \cite{wan2.1} have also been released.

In this report, we introduce Step-Video-TI2V, a new state-of-the-art TI2V model. Instead of training it from scratch, we continue the pre-training of the recently released Step-Video-T2V \cite{stepvideo}, adapting it for the image-to-video generation task. 
To evaluate the performance of different TI2V models, we build Step-Video-TI2V-Eval as a new benchmark dataset for the TI2V task. We compare Step-Video-TI2V with open-source and commercial TI2V engines on the benchmark, demonstrating its state-of-the-art performance in image-to-video generation.

Our contributions are four-fold. First, Step-Video-TI2V is a powerful open-source TI2V model with the largest model size to date. Second, it enables explicit control over the motion dynamics of generated videos, providing users with greater flexibility. Third, it excels in the anime-style TI2V task due to the training data composition. Fourth, we introduce a new benchmark dataset specifically designed for the TI2V task, fostering future research and evaluation in this domain.
\section{Model}

\begin{figure*}[t]
    \centering
    \includegraphics[width=1.0\textwidth]{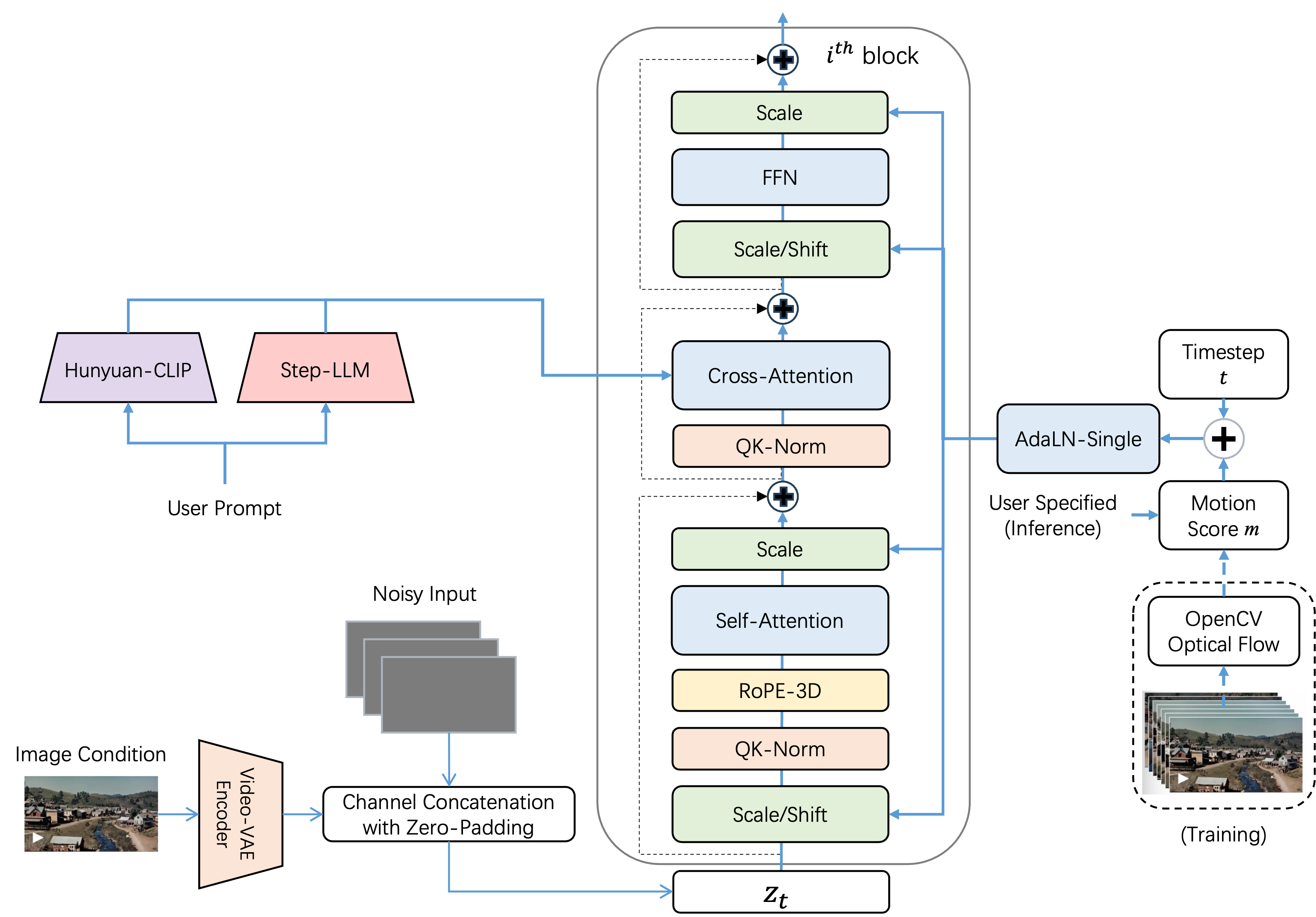}
    \caption{Overview of Step-Video-TI2V. Based on the pre-trained T2V model, we introduce two key modifications: Image Conditioning and Motion Conditioning. These enhancements enable video generation from a given image while allowing users to adjust the dynamic level of the output video.}
    \label{fig:model_overview}
\end{figure*}

\subsection{TI2V Framework}
Rather than pre-training from scratch, we train Step-Video-TI2V based on Step-Video-T2V \cite{stepvideo}, a 30B-parameter open-source text-to-video model. 
To incorporate the image condition as the first frame of the generated video, we encode it into latent representations using Step-Video-T2V’s Video-VAE and concatenate them along the channel dimension of the video latent (\S\ref{imagecondition}). 
Additionally, we introduce a motion score condition, enabling users to control the dynamic level of the video generated from the image condition (\S\ref{motioncondition}).
\Cref{fig:model_overview} shows an overview of our framework, highlighting these two modifications to the pre-trained T2V model. The details of Image Conditioning and Motion Conditioning are elaborated in \S\ref{imagecondition} and \S\ref{motioncondition}, respectively.



\subsection{Image Conditioning}\label{imagecondition}

Given an image condition of shape $[1, 3, h, w]$ (representing the number of frames, channels, height, and width), we first add random Gaussian noise to obtain a noise-augmented conditional image ~\cite{zhang2023towards,blattmann2023stable} and project it into latent space representation $\mathbf{Z}_{c}\in \mathbb{R}^{1\times c\times h'\times w'}$ using the Video-VAE in Step-Video-T2V.

We then concatenate $\mathbf{Z}_{c}$ along the channel dimension of the video latent $\mathbf{Z}_{v} \in \mathbb{R}^{f'\times c\times h'\times w'}$. To ensure compatibility with the video latent frames, we apply zero padding of shape $[f'-1, c, h', w']$ to $\mathbf{Z}_{c}$. The final latent input of DiT is ${\mathbf{Z}_{t}=[\mathbf{Z}_{v};\mathbf{Z}_{c}]\in \mathbb{R}^{f'\times 2c\times h'\times w'}}$. Thus, the channel of the corresponding patch embedding module in DiT is extended to $2c$. 



\subsection{Motion Conditioning}\label{motioncondition}
Existing open-source TI2V models generate videos directly from input text and images. However, most video generation models encounter stability issues—some generated videos exhibit high motion dynamics but contain artifacts, while others are more stable but have low motion dynamics. To balance motion dynamics and stability while providing users with control, we introduce motion embedding, enabling explicit control over the motion dynamics of the generated videos.

To incorporate motion scores into training, we use OpenCV \cite{opencv_library} to compute optical flow between video frames. Frames are sampled at 12-frame intervals, converted to grayscale, and their flow magnitudes are extracted. The highest values are selected, and the mean magnitude of the selected values is used to generate motion embeddings. In the adaptive layer normalization (AdaLN-Single) process, an additional conditional MLP block embeds the motion score, which is then combined with the embedded timestep. During inference, the motion score is set by the user as a hyperparameter, allowing direct control over the level of motion dynamics.



\section{Benchmark}

\subsection{Dataset}
\renewcommand{\arraystretch}{1.5} 

\begin{table}[ht]\small
    \centering
    \begin{tabularx}{\textwidth}{c|X}
        \toprule
        \textbf{Category} & \textbf{Fine-grained Types and Statistics} \\
        \midrule
        \multirow{3}{*}{\textbf{Real-world}} & \textbf{I2V Subject} (62), \textbf{I2V Background} (14), and \textbf{Camera Motion} (11) \\
        & \textbf{Dynamic and Artistic Elements}: \\
        & Surreal imagery (21), Musical performance (14), Artistic photography (10), Sports (39), Calligraphy and Painting (7) \\
        \midrule
        \multirow{5}{*}{\textbf{Anime-style}} & \textbf{I2V Subject} (19), \textbf{I2V Background} (25), and \textbf{Action} (13) \\
        & \textbf{Anime Style}: \\
        & 3D anime (3), Flat design (5), Illustrated characters (3), Abstract backgrounds (3), Children's drawings (5), Storybook illustrations (14), Mechanical (3), Graphic design (3), Modern aesthetics (6), Line art (2), Pixel art (2), Colored sketch (1) \\
        & \textbf{Color Aesthetic}: \\
        & High saturation (4), High contrast (3), Ink neon (4), Black background \& colored lines (2) \\
        \bottomrule
    \end{tabularx}
    \caption{Category and Statistics of Step-Video-TI2V-Eval. The corresponding number of instances is indicated in parentheses for each category.}
    \label{tab:real_anime}
\end{table}

We build Step-Video-TI2V-Eval, a new benchmark designed for the text-driven image-to-video generation task. The dataset comprises 178 real-world and 120 anime-style prompt-image pairs, ensuring broad coverage of diverse user scenarios. To achieve comprehensive representation, we developed a fine-grained schema for data collection in both categories.


Expanding on prior work \cite{huang2024vbenchcomprehensiveversatilebenchmark}, which categorized \textbf{I2V Subject}, \textbf{I2V Background}, \textbf{Action} and \textbf{Camera Motion}, we further structured the dataset based on category-specific attributes. For real-world scenes, we incorporated \textbf{Dynamic and Artistic Elements} (e.g., surreal imagery and musical performances). For anime-style scenes, we categorized samples by \textbf{Anime Style} (e.g., 3D anime, children's illustrations) and \textbf{Color Aesthetic} (e.g., ink-based designs, high saturation). A detailed dataset distribution is provided in Table \ref{tab:real_anime}.

To ensure high-quality prompts, human annotators crafted test descriptions for each image, specifying \textbf{object motion} and \textbf{camera movements} expected in the generated videos.



\subsection{Evaluation Metric}


We designed three evaluation dimensions to compare the performance of two given TI2V models. Annotators are instructed to label the results as "Win-Tie-Loss" for each dimension.

\begin{itemize}[left=0cm]
\item \textbf{Instruction Adherence}: This dimension assesses whether the models generate actions, camera movements, objects, expressions, and effects as specified in the input instructions. Penalties should be applied if the generated video contains incorrect camera movements, fails to execute the instructed motion or exhibits only minimal movement (e.g., remains static or resembles a slideshow), or omits elements described in the instructions. If neither model is clearly superior, a "Tie" should be assigned.

\item \textbf{Subject and Background Consistency}: This dimension evaluates whether the generated videos maintain consistency with the input image in terms of subject and background. Penalties should be applied if the subject or background undergoes significant style changes, if scene transitions occur without being specified in the input instructions, or if facial consistency varies significantly. If neither model is clearly superior, a "Tie" should be assigned.

\item \textbf{Adherence to Physical Laws}: This dimension examines how well the models handle multi-object interactions, motion realism, lighting, and collisions. Penalties should be applied if elements such as human bodies, animals, objects, or backgrounds exhibit distortions or unnatural deformations. If neither model is clearly superior, a "Tie" should be assigned.
\end{itemize}

\section{Experiments}

\subsection{Training Data}
We constructed a TI2V dataset comprising 5M text-image-video triples and continued training Step-Video-T2V based on it. This dataset was carefully filtered to ensure balanced motion dynamics, strong visual aesthetics, diverse concepts, accurate captions, and seamless scene continuity. 
Due to the initial design of the model, \textbf{over $80\%$ of the training data consists of anime-style videos} and \textbf{only anime-style data were used in the first half of the training stage}. These factors significantly enhance our model's performance in this category but simultaneously reduce its effectiveness on real-world-style videos (see results list in Table \ref{ranking}). 

To enable motion control, we extracted motion scores for all videos in the training data and set thresholds to filter out videos with excessively high or low motion. We integrated the motion score into our model, as described in \S\ref{motioncondition}. Additionally, we fine-tuned our in-house video captioning model for the TI2V task, focusing primarily on \textbf{describing the motion dynamics of objects and camera movements}. For example, captions like "\textit{a flock of birds flying over a tree at sunset, camera pans left}" provide detailed motion context to the TI2V task.

\subsection{Evaluation Result}

\subsubsection{Evaluation on Step-Video-TI2V-Eval}
We compare Step-Video-TI2V with two recently released open-source TI2V models, OSTopA and OSTopB, as well as two close-source commercial TI2V models, CSTopC and CSTopD. All these four models originate from China.

\begin{table}[ht]\scriptsize
\centering
\begin{tabular}{c|c|c|c|c|c}
\hline
Evaluation Dimension & Domain & vs. OSTopA & vs. OSTopB & vs. CSTopC & vs. CSTopD \\
\hline
\hline

\multirow{2}{*}{Instruction Adherence} & Real & 37-63-79 & \cellcolor{green!20}{101-48-29} & 41-46-73 & \cellcolor{green!20}{92-51-18}\\
& Anime & 40-35-44 & \cellcolor{green!20}{94-16-10} & \cellcolor{green!20}{52-35-47} & \cellcolor{green!20}{87-18-17}\\
\hline

\multirow{2}{*}{Subject and Background Consistency} & Real & \cellcolor{green!20}{46-92-39} & 43-71-64 & 45-65-50 & 36-77-47 \\
& Anime & \cellcolor{green!20}{42-61-18} & \cellcolor{green!20}{50-35-35} & 29-62-43 & \cellcolor{green!20}{37-63-23} \\
\hline

\multirow{2}{*}{Adherence to Physical Laws} & Real & \cellcolor{green!20}{52-57-49} & \cellcolor{green!20}{71-40-66} & 58-33-69 & \cellcolor{green!20}{67-33-60} \\
& Anime & \cellcolor{green!20}{75-17-28} & \cellcolor{green!20}{67-30-24} & \cellcolor{green!20}{78-17-39} & \cellcolor{green!20}{68-41-14} \\
\hline
\hline

Total Score & & \cellcolor{green!20}{292-325-277} & \cellcolor{green!20}{426-240-228} & 303-258-321 & \cellcolor{green!20}{387-283-179} \\
\hline
\end{tabular}
\caption{Comparison with baseline TI2V models using Step-Video-TI2V-Eval. 4 prompts were rejected by CSTopC. 15 prompts were rejected by CSTopD. }
\label{ranking}
\end{table}

Generally, Step-Video-TI2V achieves state-of-the-art performance on the total score across the three evaluation dimensions in Step-Video-TI2V-Eval, either outperforming or matching leading open-source and closed-source commercial TI2V models.

Specially, Step-Video-TI2V performs worse in the Instruction Adherence dimension compared to OSTopA and CSTopC, primarily due to the limited inclusion of real-world-style training data during the training phase. We will continue refining Step-Video-TI2V with more balanced training data and release an improved checkpoint in the near future. We also observed that Step-Video-TI2V performs exceptionally well on test cases with camera movement requirements, thanks to the strong performance of our video captioning model in accurately describing camera movements in videos.

\subsubsection{Evaluation on VBench-I2V}

VBench \cite{huang2024vbenchcomprehensiveversatilebenchmark} is a comprehensive benchmark suite that deconstructs “video generation quality” into specific, hierarchical, and disentangled dimensions, each with tailored prompts and evaluation methods. We utilize the VBench-I2V benchmark to assess the performance of Step-Video-TI2V alongside other TI2V models. 

Evaluation results are listed in Table \ref{tab:evaluate_vbench}, which show that Step-Video-TI2V achieves state-of-the-art results on VBench-I2V, comparing to the two latest released TI2V models. Due to the simpler motion complexity of VBench-I2V compared to Step-Video-TI2V, the issue of instruction adherence is less pronounced in this benchmark. We also presented two results of Step-Video-TI2V, with the motion set to 5 and 10, respectively. As expected, this mechanism effectively balances the motion dynamics and stability (or consistency) of the generated videos.

\begin{table}[ht]\scriptsize
\centering
\begin{tabular}{c|c|c|c|c}
\hline
Scores & Step-Video-TI2V (motion=10) & Step-Video-TI2V (motion=5) & OSTopA & OSTopB \\
\hline
\hline
Total Score & \cellcolor{green!20}{87.98} & 87.80 & 87.49 & 86.77 \\
\hline
\hline
I2V Score & 95.11 & \cellcolor{green!20}{95.50} & 94.63 & 93.25 \\
\hline
Video-Text Camera Motion & 48.15 & \cellcolor{green!20}{49.22} & 29.58 & 46.45 \\
\hline
Video-Image Subject Consistency & 97.44 & \cellcolor{green!20}{97.85} & 97.73 & 95.88 \\
\hline
Video-Image Background Consistency & 98.45 & 98.63 & \cellcolor{green!20}{98.83} & 96.47 \\
\hline
\hline
Quality Score & \cellcolor{green!20}{80.86} & 80.11 & 80.36 & 80.28 \\
\hline
Subject Consistency & 95.62 & 96.02 & 94.52 & \cellcolor{green!20}{96.28} \\
\hline
Background Consistency & 96.92 & 97.06 & 96.47 & \cellcolor{green!20}{97.38} \\
\hline
Motion Smoothness & 99.08 & \cellcolor{green!20}{99.24} & 98.09 & 99.10 \\
\hline
Dynamic Degree & 48.78 & 36.58 & \cellcolor{green!20}{53.41} & 38.13 \\
\hline
Aesthetic Quality & 61.74 & \cellcolor{green!20}{62.29} & 61.04 & 61.82 \\
\hline
Imaging Quality & 70.17 & 70.43 & \cellcolor{green!20}{71.12} & 70.82 \\
\hline
\end{tabular}
\caption{Comparison with two open-source TI2V models using VBench-I2V.}
\label{tab:evaluate_vbench}
\end{table}

\section{Conclusion}
In this work, we introduce Step-Video-TI2V, an open-source model that achieves state-of-the-art performance on the TI2V task. Our model offers explicit control over motion dynamics, providing users with enhanced flexibility in video generation. It also excels in the anime-style TI2V task, owing to the specific composition of the training data. To advance the field further, we present a new benchmark dataset tailored for the TI2V task, laying the groundwork for future research and evaluation. We hope our contributions will drive innovation in video generation and empower the broader research and application communities.

{
\small
\bibliographystyle{unsrtnat}
\bibliography{main}
}

\newpage
\section*{Contributors and Acknowledgments}\label{team}

We designate core contributors as those who have been involved in the development of Step-Video-T2V throughout its entire process, while contributors are those who worked on the early versions or contributed part-time. Contributors are listed in \textbf{alphabetical order by first name}.

\begin{itemize}[left=0cm] 
\item \textbf{Core Contributors:}
    \begin{itemize}
        \item \textbf{Model \& Training:}
        Haoyang Huang, Guoqing Ma, Nan Duan.
        \item \textbf{Infrastructure:} 
        Xing Chen,
        Changyi Wan, Ranchen Ming.
        \item \textbf{Data \& Evaluation:} Tianyu Wang, Bo Wang, Zhiying Lu,  Aojie Li, Xianfang Zeng, Xinhao Zhang, Gang Yu,  Yuhe Yin, Qiling Wu, Wen Sun, Kang An, Xin Han, Deshan Sun, Wei Ji.
    \end{itemize}

\item \textbf{Contributors:} 
Bizhu Huang, Brian Li, Chenfei Wu, Guanzhe Huang, Huixin Xiong, Jiaxin He, Jianchang Wu, Jianlong Yuan, Jie Wu, Jiashuai Liu, Junjing Guo, Kaijun Tan, Liangyu Chen, Qiaohui Chen, Ran Sun, Shanshan Yuan, Shengming Yin, Sitong Liu, Wei Chen, Yaqi Dai, Yuchu Luo, Zheng Ge, Zhisheng Guan, Xiaoniu Song, Yu Zhou.

\item \textbf{Project Sponsors:} 
Binxing Jiao, Daxin Jiang, Heung-Yeung Shum, Jiansheng Chen, Jing Li, Shuchang Zhou, Xiangyu Zhang, Xinhao Zhang, Yi Xiu, Yibo Zhu.

\item \textbf{Corresponding Authors:} 
Daxin Jiang (djiang@stepfun.com), Nan Duan (nduan@stepfun.com).
\end{itemize}

\end{document}